\documentclass[letterpaper, 10 pt, conference]{ieeeconf}  

\usepackage{algorithmic}
\usepackage{epsfig}
\usepackage{graphicx}
\usepackage{amsmath}
\usepackage{amssymb}
\usepackage{booktabs}
\usepackage{multirow}
\usepackage{multicol}
\usepackage[ruled,vlined]{algorithm2e}
\usepackage{xcolor}
\usepackage{pifont}
\usepackage{xcolor}
\usepackage{csquotes}
\IEEEoverridecommandlockouts                              

\usepackage{booktabs}                                   
\usepackage{multirow}                                   
\usepackage{makecell}                                   
\usepackage{tablefootnote}                              
\usepackage[symbol]{footmisc}                           
\usepackage{amsmath,amssymb}                            
\usepackage{xcolor}                                     
\let\labelindent\relax              
\usepackage{enumitem}                                   
\usepackage{subcaption}                                 
\usepackage{stfloats}                                   
\usepackage[misc]{ifsym}     
\usepackage{hyperref}   

\newcommand{\eat}[1]{}                                  

\title{\LARGE \bf
Visual Robotic Manipulation with Depth-Aware Pretraining
}

\author{Wanying Wang$^{1,*}$, Jinming Li$^{1,*}$, Yichen Zhu$^{2,*}$, Zhiyuan Xu$^{2}$, Zhengping Che$^{2}$,\\
Yaxin Peng$^{1,\dagger}$, Chaomin Shen$^{3}$, Dong Liu$^{2}$, Feifei Feng$^{2}$, and Jian Tang$^{2,\dagger}$
\thanks{$^{1}$Department of Mathematics, School of Science, Shanghai University, China
        {\tt\small \{wywang, ljm2022, yaxin.peng\}@shu.edu.cn}}
\thanks{$^{2}$Midea Group, China
        {\tt\small \{zhuyc25, xuzy70, chezp, liudong13, feifei.feng, tangjian22\}@midea.com}}
\thanks{$^{3}$School of Computer Science, East China Normal University, China
        {\tt\small cmshen@cs.ecnu.edu.cn}}
\thanks{
    $^*$Equal contributions. This work was done during Wanying Wang and Jinming Li's internship at Midea Group.
}
\thanks{
    $^\dagger$Corresponding authors: Yaxin Peng and Jian Tang.
}
}
\usepackage{pifont}
\usepackage{graphicx} 

\begin{document}

\maketitle
\thispagestyle{empty}
\pagestyle{empty}

\begin{abstract}
Recent work on visual representation learning has shown to be efficient for robotic manipulation tasks. However, most existing works pretrained the visual backbone solely on 2D images or egocentric videos, ignoring the fact that robots learn to act in 3D space, which is hard to learn from 2D observation. In this paper, we examine the effectiveness of pretraining for vision backbone with public-available large-scale 3D data to improve manipulation policy learning. Our method, namely \textit{Depth-aware Pretraining for Robotics (DPR)},  enables an RGB-only backbone to learn 3D scene representations from self-supervised contrastive learning, where depth information serves as auxiliary knowledge. No 3D information is necessary during manipulation policy learning and inference, making our model enjoy both efficiency and effectiveness in 3D space manipulation. Furthermore, we introduce a new way to inject robots' proprioception into the policy networks that makes the manipulation model robust and generalizable. We demonstrate in experiments that our proposed framework improves performance on unseen objects and visual environments for various robotics tasks on both simulated and real robots. 
\end{abstract}

\section{Introduction}
The ability of robots to interpret and interact with visual environments is crucial, especially in service-oriented tasks. Thanks to significant advancements in pretrained visual representations, robots now possess an unparalleled ability to generalize on both simulated and real-world scenarios. Such progress paves the way for robots to adapt flexibly to diverse user scenarios, ensuring robust service delivery. 

Existing works typically rely on self-supervised objectives and train on large-scale 2D image datasets or ego-centric video demonstrations to understand the world through excessive access to a vast amount of data. 
However, such an approach shows its limitations, as the robot is required to act in 3D space that 2D images typically fail to represent well. 
A workaround is to utilize 3D knowledge, depth map, point cloud, or reconstructed 3D scene~\cite{shen2023distilled, clipfields} to help the robot perform in 3D space. 
Yet, the 3D models~\cite{qi2017pointnet} are typically computationally heavy and hard to deploy on edge devices, not to mention the expansiveness of acquiring accurate depth cameras for products. 
Therefore, a natural question arises: How can we harness the potential of visual models in robotic manipulation without access to 3D information? 


In this paper, we took a different path, investigating how to leverage public-available large-scale 3D datasets to boost visual robotic manipulation via pretraining. 
We expect that, without access to depth information during policy learning or task inference, the visual backbone could capture task-relevant features to enhance manipulation performance in 3D space. 
%
Our methodology draws on the strengths of contrastive learning (CL). Distinct from conventional CL that trains on 2D image data or text-image pairs, our system utilizes depth images, treating them as a fresh reservoir of data samples for image-level positive/negative selection. Central to our method is the use of external depth data when choosing positive/negative pairings. Depth map not only provides intricate semantic insights into object boundaries but also inhibits the acquisition of ambiguous representations near these edges. 
Our empirical analysis shows that our pretraining model is concentrated on task-relevant objects. Beyond our pretrained framework, we have an effective way to integrate robot proprioception with policy networks. This is driven by the idea that the label itself could harbor critical information guiding the policy network to identify the correct trajectory. 
Our experiments show that our proposed framework significantly enhances visual robotic manipulation in various tasks, both in simulation and real-world scenarios. It is also noteworthy that our method is a versatile plug-and-play module, readily adaptable to numerous existing end-to-end manipulation models to augment inference.

\noindent
\textbf{In summary, our contributions are the following:}
\begin{itemize}
    \item We build a depth-aware pretraining framework that facilitates the visual robotic manipulation tasks without access to the depth information during both policy training and inference phases. 
    \item We propose a novel proprioception injection method 
    that extracts the useful representation in the robot state and facilitates the information fusion on the deep neural networks. 
    \item Empirical evaluations and ablations on both simulated and real robots validate the superiority of our framework over existing baselines. 
\end{itemize}

\section{Related Works}
\noindent
\textbf{Visual Representation for Robotics.} Representation learning is critical in high-dimensional control settings, particularly when managing visual observation spaces. Recent advancements in this field have explored feature representation for robotics, especially for the pretrained visual representation~\cite{yuan2022pre, hansen2022pre, parisi2022unsurprising}. Shapebias~\cite{shapebias} discusses the shape-aware features in the pretraining stage that boost the downstream manipulation tasks, where DINO~\cite{dino, oquab2023dinov2} are the most shape salient unsupervised models. RoboAdapter~\cite{sharma2023lossless} uses a parameter-efficient fine-tuning strategy to finetune the pretrained visual backbone. Supervised pretraining have been extensively explored. Bridgedata~\cite{bridgedata, walke2023bridgedata} train policies with cross-domain datasets. 
Many literature~\cite{jiang2022vima, hill2020human, jang2022bcz, brohan2022rt1, nair2022learning, lynch2020language, reed2022generalist, shridhar2023perceiver} use visual encoders from pretrained vision-language models to train the policy networks. Unlike prior works, our work seeks to use large-scale datasets with the help of depth information during pretraining and transfer such 3D representation to robotics control without further access to depth knowledge.

\noindent
\textbf{Self-Supervised Learning.} The self-supervised learning has succeeded greatly in vision~\cite{dai2022cluster} and natural language processing. There are several types of self-supervised learning, including masked auto-encoder~\cite{maskvisual}, contrastive learning~\cite{moco}, and siamese frameworks~\cite{chen2021exploring}, and world model~\cite{mendonca2023structured, seo2023multi}. There have been extensive prior works studied to use self-supervised learning as a pre-trained objective to facilitate grasp~\cite{florence2019self, kalashnikov2018qt, pinto2016supersizing, danielczuk2020exploratory, fu2022legs} or manipulation~\cite{kalashnikov2021mt}. Yet, in which what types of visual representation are beneficial to robotic manipulation is still an open question. Prior works have verified the practical approach using masked auto-encoder~\cite{mae} on robotics manipulation in both simulated and real environments~\cite{xiao2022masked, maskvisual, radosavovic2023robot}. Pri3D~\cite{pri3d} draws from the foundational idea of employing prior 3D data to augment contrastive learning. This work investigates contrastive learning-based approaches that use depth maps to enhance the pretrained visual representation in 3D space to enhance manipulation performance. 
\begin{figure*}
\centering
\includegraphics[width=0.9\linewidth]{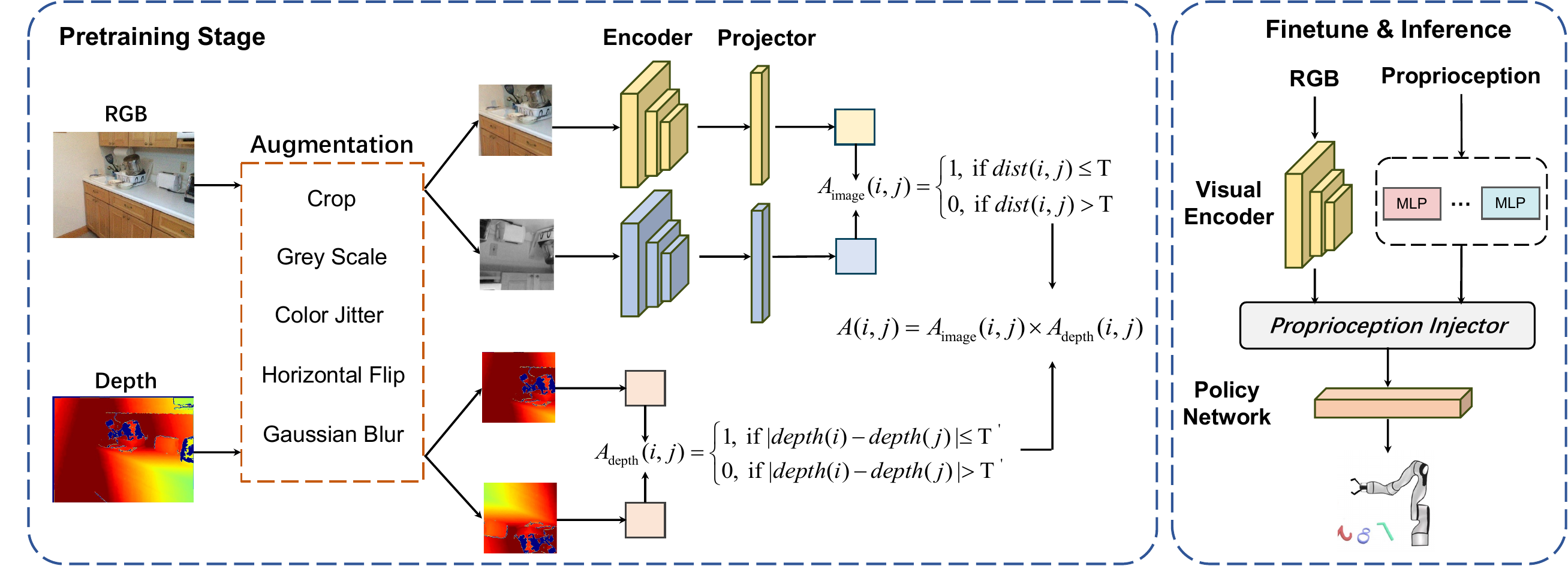}
\caption{An overview of our depth-aware pretraining framework for robotics. We apply identical augmentations to both the input RGB image and the depth map. The two resulting cropped images are channeled through the encoder and the projector. Subsequently, the distance between the two feature maps is computed to distinguish between positive and negative pairs. The pair of depth crops is resized to match the shape of the feature map, and the depth discrepancy is considered. The final decision for positive or negative pairs is based on a combined assessment of both RGB features and depth. During the inference phase, the pretrained encoder is employed for subsequent robotic manipulation tasks.}
\label{fig:pretrain}
\end{figure*}

\section{Method}
In this section, we provide a thorough discussion of our proposed depth-aware pretraining methods. Then, we present our proprioception injection method. 

\subsection{Depth-aware Self-Supervised Learning}
\subsubsection{Pixel-Level Contrastive Learning}
Given an input RGB image and its corresponding depth map, we initially crop the RGB image into two views and introduce augmentations to these crops, following the typical procedure in contrastive learning settings. The two crops are then resized to a consistent resolution and processed through both an encoder and a momentum encoder. The resulting feature maps are subsequently projected using a projector to determine the distance between the two crops. The positive/negative pairs in the image level are distinguished by 
\begin{align}
A_{\text {image }}(i, j)= \begin{cases}1, & \text { if } \operatorname{dist}(i, j) \leq \mathcal{T} \\ 0, & \text { if } \operatorname{dist}(i, j)>\mathcal{T}\end{cases}
\end{align}
where $i$ and $j$ are the indexes of vectors in the first and second view, respectively; $\operatorname{dist}$ denotes the normalized euclidean distance between $i$ and $j$ in feature space; $\mathcal{T}$ is the distance threshold.

For the depth map, we crop it from the same position as the RGB image does and employ the shared augmentation. 
We then map the cropped depth map to the size of a 7 $\times$ 7 feature map that allows us to calculate the pairwise difference between the pixels of the resized depth crops. Then, the selection of positive/negative pairs at the depth level can be constructed as 
\begin{align}
    A_{\text {depth }}(i, j)= \begin{cases}1, & \text { if }|\operatorname{depth}(i)-\operatorname{depth}(j)| \leq \mathcal{T}^{\prime} \\ 0, & \text { if }|\operatorname{depth}(i)-\operatorname{depth}(j)|>\mathcal{T}^{\prime}\end{cases}
\end{align}
where $\operatorname{depth}$ is the normalized depth value and $\mathcal{T}^{\prime}$ is the threshold on the depth maps. In our implementation, we set both $\mathcal{T}$ and $\mathcal{T}^{'} $ as $\{0.3, 0.5, 0.7\}$ followed \cite{pixdepth}, and average all the contributions in the loss function. Now the final positive/negative pairs selection is obtained by element-wise multiplication of both $A_{\text {image }}(i, j)$ and $A_{\text {depth }}(i, j)$, i.e.,
\begin{align}
A(i, j)=A_{\text {image }}(i, j) \times A_{\text {depth }}(i, j).
\end{align}
Based on the leveraged depth information, for the first-view pixel $i$ that is also located in the second view, the positive and negative groups of pixels are defined as
\begin{align}
\Omega_p^i= \{ j\in \Omega^j, A(i, j)=1 \}, \\
\Omega_n^i= \{ j\in \Omega^j, A(i, j)=0 \},
\end{align}
where $\Omega^j$ is the group of pixels in the second view. Then the  contrastive learning objective $\mathcal{L}_{pix}(i)$ is
\begin{align}
-\log \frac{
\mathop{\sum}_{j \in \Omega_p^i}
e^{\cos \left(\mathbf{x}_i, \mathbf{x}_j^{\prime}\right) / \tau}}{\sum_{j \in \Omega_p^i} e^{\cos \left(\mathbf{x}_i, \mathbf{x}_j^{\prime}\right) / \tau}+\sum_{k \in \Omega_n^i} e^{\cos \left(\mathbf{x}_i, \mathbf{x}_k^{\prime}\right) / \tau}},
\end{align}
where
$\mathbf{x}_i$ and $\mathbf{x}_j^{\prime}$ are the pixel feature vectors in two views, and $\tau$ is a scalar temperature hyper-parameter, set by default to 0.06. For the first view, all pixels in the intersection of both views are averaged to compute the loss. In a parallel manner, the second view calculates contrastive loss $\mathcal{L}_{pix}(j)$ and averages for each pixel denoted by $j$. The total pixel-level contrastive learning loss is constructed as
\begin{align}
\mathcal{L}_{pix}=\frac{1}{\sum_{i}\sum_{j}}\sum_{i}\sum_{j}\left(\mathcal{L}_{pix}(i)+\mathcal{L}_{pix}(j)\right).
\label{eq:pix}
\end{align}

\begin{figure}[t]
\centering
\includegraphics[width=0.95\linewidth]{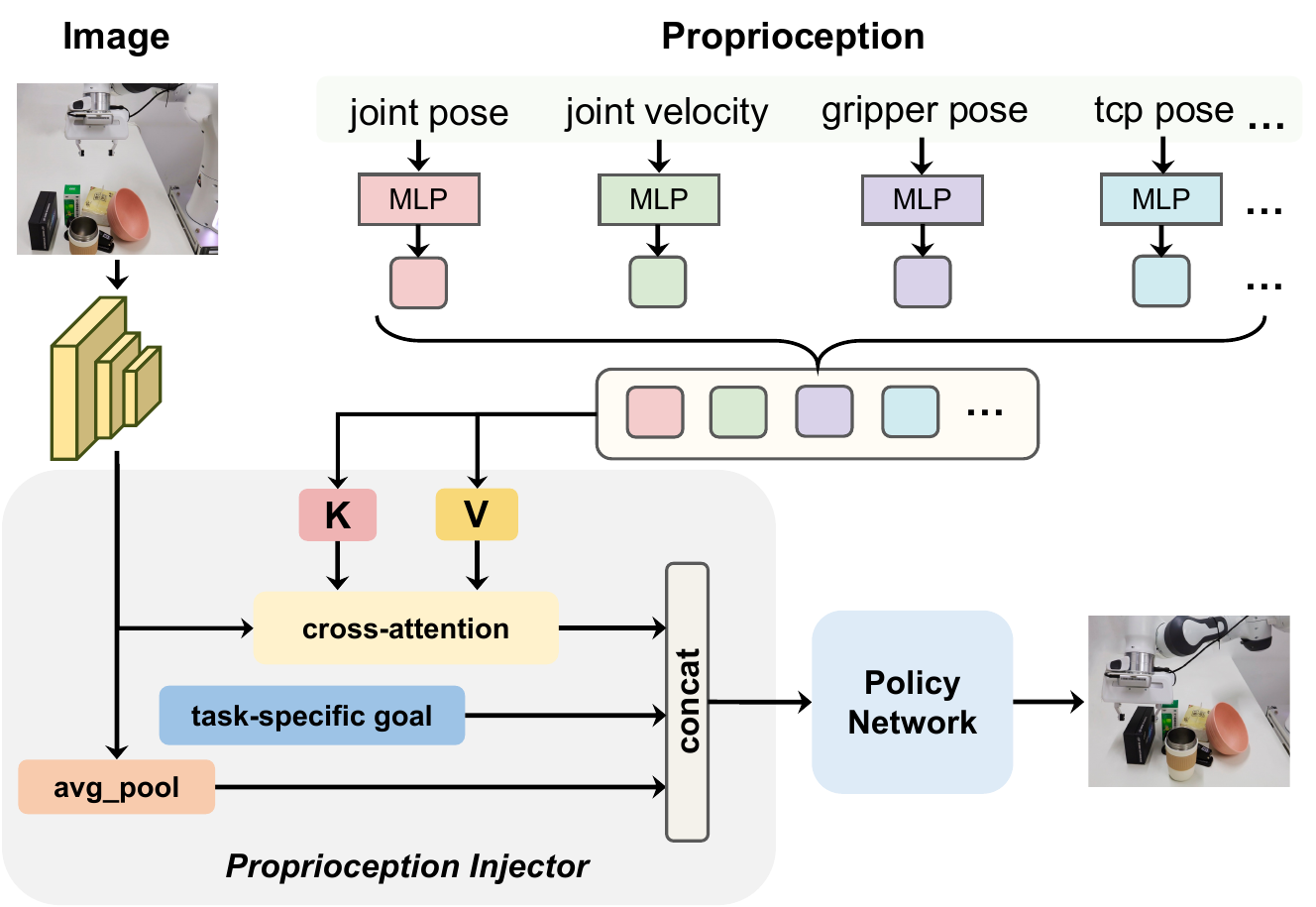}
\caption{Architecture of the proprioception injection method. }
\label{fig:method}
\end{figure}
\subsubsection{Combined with Instance-level Objective}
Our approach doesn't exclusively rely on pixel-level contrastive learning, given that robotic manipulation isn't particularly sensitive to pixel-level features. As such, the described pixel-level training process is amalgamated with instance-level pretext tasks, utilizing the same data loader and backbone encoders. Specifically, an independent projection head is applied to the output of the encoder, producing $\mathbf{q}$ and $\mathbf{q'}$ for the two views, respectively. Further, a prediction head is followed after the projector, producing $\mathbf{k}$ and $\mathbf{k'}$. In this implementation, the instance-level learning objective is constructed as
\begin{align}
\mathcal{L}_{ins}=(\mathbf{q}^{T}\mathbf{k'} +\mathbf{k}^{T}\mathbf{q'})/2.
\label{eq:ins}
\end{align}
The two losses from the pixel-level \eqref{eq:pix} and instance-level \eqref{eq:ins} tasks are balanced by a multiplicative factor $\alpha$ (set to 1):
\begin{align}
\mathcal{L}=\mathcal{L}_{pix}+\alpha\mathcal{L}_{ins}.
\end{align}

In summary, the mask $A$ identifies clusters of positive/negative pixels, which are then incorporated into the loss equation. To distill the essence, this method seamlessly integrates depth prior data into the learning process, eliminating the encoder's need to access depth maps during the training phase. Concurrently, the delineation draws from the semantic insights offered by depth maps rather than merely relying on $\operatorname{dist}(i, j)$ within a projected 2D image. Such a method could inadvertently amalgamate distinct objects. As visualized in Figure \ref{fig:pretrain}, the pretrained visual encoder is integrated into the inference stage.

\noindent
\textbf{Resolution Adaptation.} Increasing image resolution plays an important role in robotic manipulation, particularly when object-aware representations are essential for accurate task performance. This becomes even more crucial as small objects, pivotal for certain tasks, might fade or become indistinguishable at lower resolutions, thus complicating the challenges faced in robotics. Yet, the computational overhead of training high-resolution input, in terms of both time and memory, spikes considerably, particularly when one considers the vast sizes of the pretraining datasets that are typically employed \cite{oquab2023dinov2}. To address these challenges, we propose a resolution adaptation strategy.  By initiating training with a low resolution during the initial 90\% of the training epochs, we can maintain efficiency and speed. Then, as we approach the final stages of training, the resolution is amped up for the remaining 10\% of epochs. This tactic ensures that while the bulk of the training process is more manageable and faster, the final stages still benefit from the granularity and detail of higher-resolution images, thus harnessing the best of both worlds. Surprisingly, we observe this method also prevents overfitting in the pretraining stage, which has long been known to be beneficial for the downstream tasks. 
\noindent

\begin{table}[t]
\renewcommand{\arraystretch}{1.3}
\caption{An example of proprioception in PickCube task. }
\centering
\begin{tabular}{ll}
\toprule
Proprioception & Numerical Value \\
\midrule
Joint Position & [-0.1421, 0.7487, 0.0087, \\
&-1.7940, -0.0147, 2.5699, 0.5372]\\
Gripper Position & [0.0354, 0.0354]\\
Joint Velocity & [-0.1196, 0.2343, 0.0016, \\
&0.1463, 0.0020, 0.0711, -0.1380]\\
Gripper Velocity & [-0.0113, -0.0116]\\
TCP Position & [0.0330, -0.0873, 0.0298]\\
TCP Rotation & [0.0037, 0.9980, 0.0622, 0.0132]\\
Goal Position & [0.0495, 0.0291, 0.0847]\\
TCP to Goal Position & [0.0165, 0.1164, 0.0550]\\
\bottomrule
\end{tabular}
\label{table:proprioception}
\end{table}

\subsection{Proprioception Injection Method}
The robot’s proprioception is critical knowledge that helps the robot to have an understanding of its previous state, which can potentially be helpful to the generalization of the robot. Designing an appropriate method for injecting proprioception poses two primary challenges. Firstly, how do we distill valuable information from proprioception? Notably, proprioceptive data can vary between robots, and there can be discrepancies between simulated and real robots. The second challenge stems from the fact that different proprioceptions represent completely different states of the robot. An illustrative example of proprioception values can be seen in Table~\ref{table:proprioception}, where the Tool Center Point (TCP) is the center point between the gripper’s two fingers. Merely concatenating these data and employing a single linear layer for processing could undermine the model's trainability. As a result, we propose a way to inject proprioception into the training process. 

We introduce a new type of alignment technique that helps integrate the robot’s proprioception into the policy network (Figure \ref{fig:method}). Specifically, denoting the proprioception input as a set of different states $S = \{s_{1}, s_{2}, \cdots, s_{n}\}$, each $s_{i}$ represents a state such as joint velocity, joint position, and etc. We use a set of linear layers, denoted as $Linear = \{l_{1}, l_{2}, \cdots, l_{n}\}$, where each $l_{i}$ with a dimension of [$\dim(s_{i})$,
256, 8]. Then we obtain the corresponding output of $O = \{l_{1}(s_{1}), l_{2}(s_{2}), \cdots, l_{n}(s_{n})\}$. The mapped state is concatenated together, resulting in a sequence of size $n \times 8$. A layer normalization is followed after $O$ to normalize the numerical values that come from different robot states.

To align the proprioception features with the image feature, we adopt a cross-attention \cite{vaswani2017attention} to connect these two parts. Given a pretrained visual representation $\mathcal{F}$, the input RGB image $I$ and the proprioception $S$, we have the visual embedding $z=\mathcal{F}(I)$ and the processed proprioception $O=Linear(S)$. The cross-attention layer then generates 
\begin{align}
\operatorname{Attn}(z, O)=\operatorname{Softmax}\left(\frac{Q_{z}\left(K_{O}\right)^T}{\sqrt{d}}\right) V_{O},
\end{align}
where $Q_{z}$ is adapted visual embedding $z$, $K_{O}$ and $V_{O}$ are from the proprioception $O$, $d$ is the embedding dimension.

\noindent
\textbf{Training Objective.} We adopt the behavior cloning (BC) loss
\begin{align}
L_{BC}=\|\pi(z,\operatorname{Attn}(z, O), goal)-a\|^2
\end{align}
to train policy $\pi$, where $goal$ is the task-specific goal information, $a$ is the action in the expert data. We parameterize $\pi$ as a three-layer MLP, with the concatenation of AvgPooling-processed $z$, $\operatorname{Attn}(z, O)$ and $goal$ as the input.

\begin{figure}[t]
\centering
\includegraphics[width=0.8\linewidth]{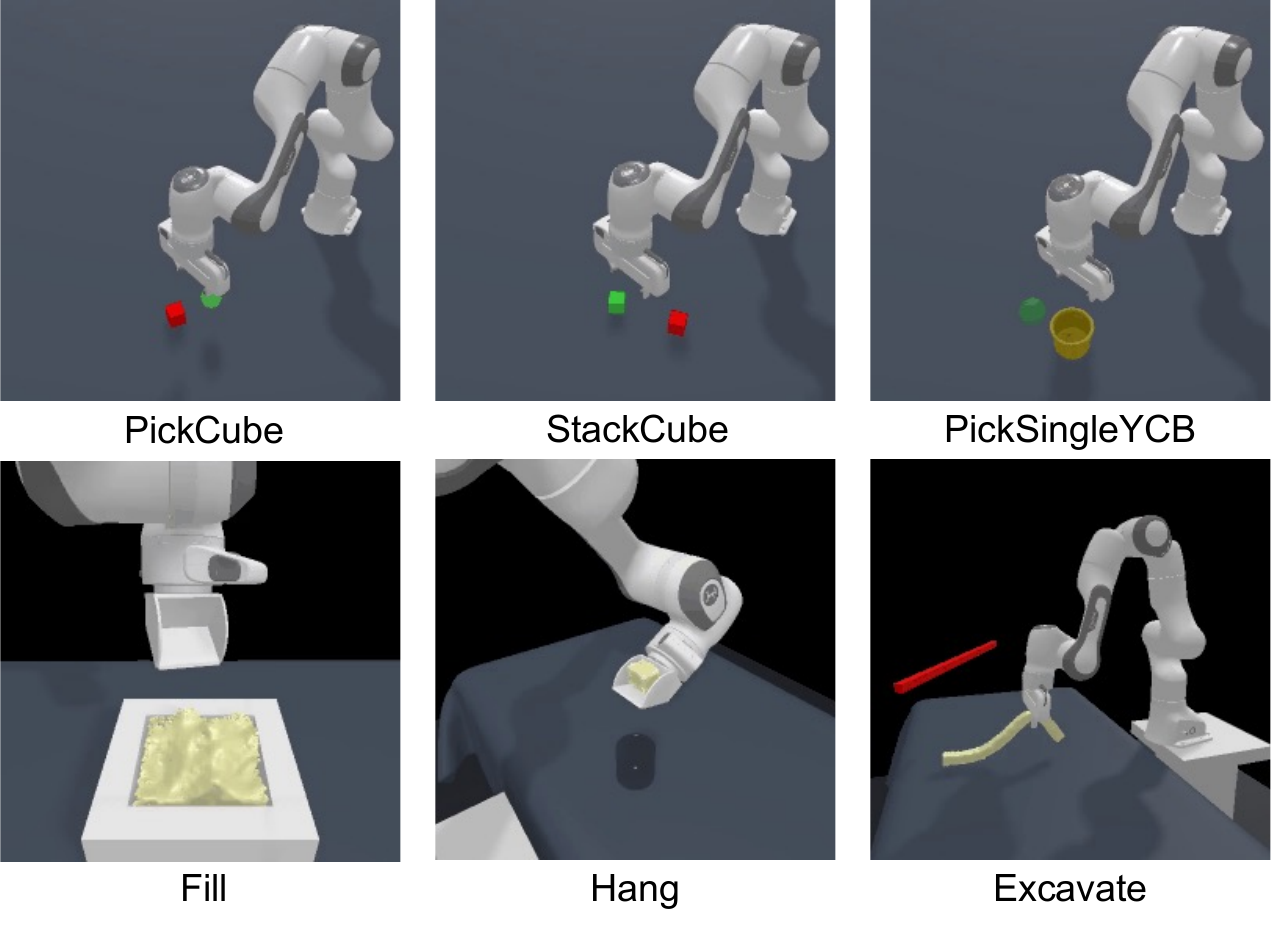}
\caption{We use six contact-rich object manipulation tasks from ManiSkill2 for simulation. Top: rigid-body tasks. Bottom: soft-body tasks.}
\label{fig:task}
\end{figure}

\section{Experiment}
\begin{table*}[htbp]
\renewcommand{\arraystretch}{1.3}
\caption{Success rates on rigid body and soft body tasks. Our method consistently outperforms Baseline and SOTA in all six environments. All metrics are reported in percentage $(\%)$ with the best ones bolded. }
\centering
\begin{tabular}{lccccccc}
\toprule
Methods &  Dataset 
&PickCube & StackCube & PickSingleYCB & Fill & Hang & Excavate  \\
\midrule
\multicolumn{8}{c}{RGB Pretraining} \\
\midrule
Train-from-scratch & -&45 & 91.3&  35.7& 42.5 & 83.2& 20.1\\
Supervision~\cite{he2016deep}  & ImageNet&46.3 & 92& 27.6 & 61.3 & 88.4&19  \\
MOCO~\cite{moco}  & ImageNet&71.3 & 91.3&  40.7& 46.9 & 82.7& 14\\
PixPro~\cite{pixpro} & ImageNet & 9.4 & 90.1 & 27.5 & 67.5 & 88.8&13.6  \\
\midrule
\multicolumn{8}{c}{RGBD Pretraining} \\
\midrule
RGBD Fusion & -& 60.1 & 90.1& 28.2 & 48.8 & 85.7& 19.4 \\
Pri3D~\cite{pri3d} & ScanNet&10.7 & 90& 19.4& 57.4 & 88.4& 17\\
Ours  & ScanNet& \textbf{94.4} &\textbf{92.6} &\textbf{43.8} & \textbf{73.2}&\textbf{88.8} &\textbf{21.3} \\
\bottomrule
\end{tabular}
\label{table:main}
\end{table*}

In this section, we first introduce the experimental setup, including the tasks and implementation details. Then, we conduct quantitative evaluations in both simulation and real-world environments to validate our approach.

\subsection{Simulation Experiments}
\noindent
\textbf{Tasks.} To assess our approach, we conducted experiments using ManiSkill2~\cite{gu2023maniskill2}. As depicted in Figure~\ref{fig:task}, ManiSkill2 comprises six tasks. These include three rigid-body and three soft-body tasks, collectively representing a diverse array of manipulation behaviors that integrate both prehensile and non-prehensile motions.
\begin{itemize}
    \item \textit{PickCube}: Pick up a cube and move it to a goal position;
    \item \textit{StackCube}: Pick up a cube and place it on another cube;
    \item \textit{PickSingleYCB}: Pick and place an object from YCB;
    \item \textit{Fill}: Fill clay from a bucket into the target beaker;
    \item \textit{Hang}: Hang a noodle on the target rod;
    \item \textit{Excavate}: Scoop up a specific amount of clay and lift it to a target height.
\end{itemize}
In all tasks, the agent receives image observations alongside robot proprioception data, which includes joint positions, joint velocities, and more. We observed that the success rate of BC, as reported in the original ManiSkill2~\cite{gu2023maniskill2}, is extremely low, hindering users from making meaningful comparisons. Consequently, we collect new expert demonstrations and evaluate all methods using these newly acquired data. For rigid-body tasks, expert data is generated by training a DAPG~\cite{rajeswaran2017learning} agent on 1,000 successful trajectories (Except for PickSingleYCB, which is trained with 6,500 trajectories due to low success rates when training on 1,000 trajectories). In contrast, for soft-body tasks, we gathered 190 trajectories using task and motion planning. We then divided all the data into training, validation, and testing sets, maintaining an 8/1/1 ratio. A summary of our data collection is demonstrated in Table~\ref{tbl:data_collection}. All models are trained two times, and we report the average results over two trials.


\begin{figure*}[t]
\centering
\includegraphics[width=0.75\linewidth]{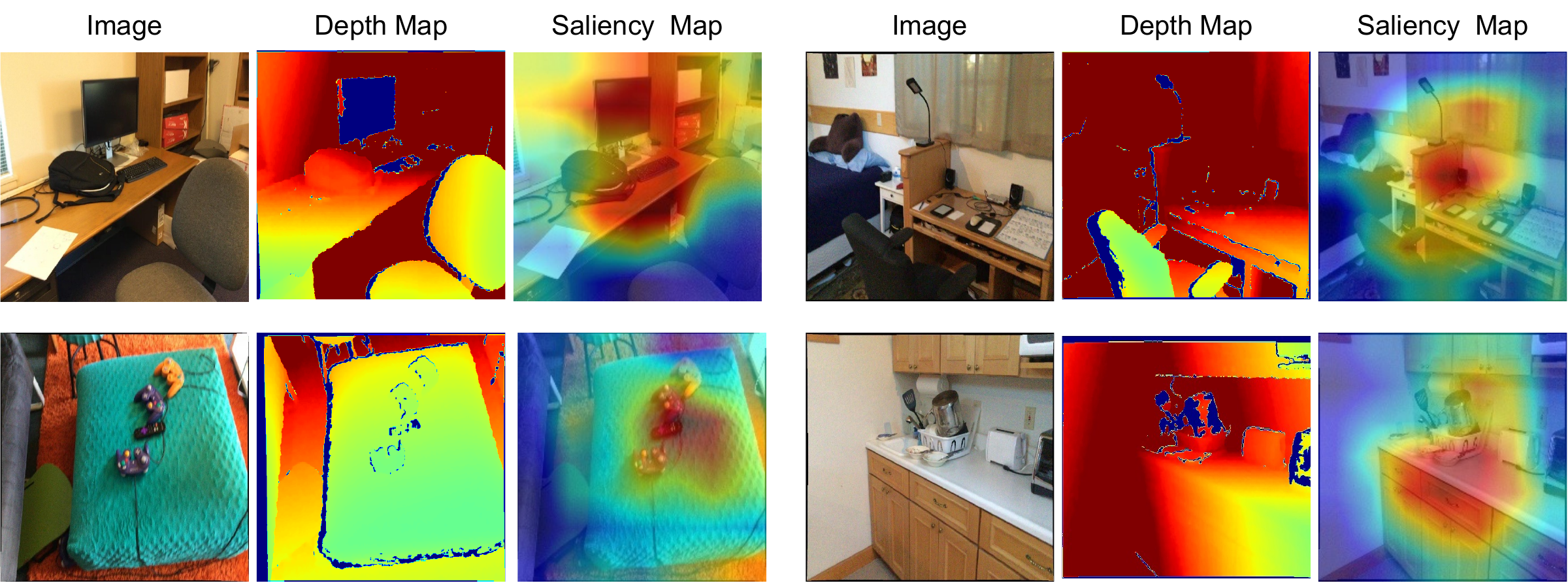}
\caption{We visualize the final embedding from depth-aware pretrained ResNet18 via GradCAM~\cite{selvaraju2017gradcam}. It appears that our pretrained model segments actionable parts for robots of the scene.}
\label{fig:visualization}
\end{figure*}
\noindent
\textbf{Baselines.}
We evaluate our model against two categories of pretraining algorithms: those incorporating depth information during pretraining and those that do not. For the former, we compare with ImageNet~\cite{deng2009imagenet} supervised pretraining, PixPro~\cite{pixpro}, and MOCO~\cite{moco}. For the latter, we look at Pri3D~\cite{pri3d}. In the latter method, we merge the RGB image with the depth data, resulting in an input dimension of $\text{Height} \times \text{Width} \times 4$. Additionally, we assess our model against training-from-scratch approaches and direct RGBD fusion.

\noindent
\textbf{Architecture Details.} In our experiments, we chose ResNet18 \cite{he2016deep} as the vision encoder for all models. Two views of RGB image are concatenated as the input of the pretrained encoder, resulting in the vision feature of $512 \times 8 \times 4$ before pooling, and the shape of the resized vision feature becomes $512 \times 32$. The policy network is implemented by a three-layer MLP of size $[\dim(embedding), 256, 128, \dim(action)]$.

\noindent
\textbf{Training Details.} During the pretraining stage, all models run for 50 epochs with ScanNet \cite{dai2017scannet} using a batch size of 1024. We use the LARS optimizer~\cite{you2017large} with a weight decay of $1e^{-5}$ and a cosine learning rate scheduler with warm up for the first 5\% of total epochs. The initial learning rate is set as $3e^{-4}$ and decayed to $1e^{-5}$, and we train in bfloat16 precision. We take random image crops of resolution 112 $\times$ 112 as input and apply augmentations. Then, we increase the resolution to 224 $\times$ 224 for the last five epochs. For the fine-tuning stage, we train the agent for 400 epochs, evaluate it online in the environment every 40 epochs, and report the best success rate achieved. We use the same hyperparameters for all methods, including various baseline approaches. 

\subsection{Experimental Results}
\noindent
\textbf{Main Experiments.} We study the importance of depth-aware pretraining by comparing the effect of pretrained vision weights. Results are shown in Table \ref{table:main}. We observe a significant improvement in success rate while adopting our depth-aware pretrained weights, which is probably because robot manipulation requires 3D knowledge, and the feature of DPR contains more spatial information. Note that it achieves better performance than directly using depth-image (RGBD Fusion), which indicates the role of effective utilization of depth information.
\begin{table}[t]
\renewcommand{\arraystretch}{1.3}
\caption{Collecting expert demonstrations for behavior cloning in six environments.}
\centering
\begin{tabular}{lc}
\toprule
Environment & Demonstration \\
\midrule
PickCube&(with pretrained DAPG) 1K Trajectories\\
StackCube  &(with pretrained DAPG) 1K Trajectories\\
PickSingleYCB& (with pretrained DAPG) 6.5K Trajectories\\
\midrule
Fill&  190 Trajectories \\
Excavate & 190 Trajectories \\
Hang& 190 Trajectories \\
\bottomrule
\end{tabular}
\label{tbl:data_collection}
\end{table}

\begin{table}[t]
\renewcommand{\arraystretch}{1.3}
\caption{The proprioception inject methods notably improve the success rate on Maniskill2.}
\centering
\resizebox{0.45\textwidth}{!}{\begin{tabular}{ccccccc}
\toprule
State & Method & Success (\%)&  Method & Success (\%) & Method & Success (\%)\\
\midrule
 \ding{54} & \multirow{2}{*}{MOCO}  & 46.9 & \multirow{2}{*}{Pri3D} & 57.4 &   \multirow{2}{*}{Ours} & 64.0\\
\ding{52} &  &50.7 &  & 61.3 &  & 73.2 \\
\bottomrule
\vspace{-1cm}
\end{tabular}}
\label{table:state}
\end{table}

\noindent
\textbf{Visualization of Pretrained Vision Backbone.} An interesting artifact of training a CNN with DPR is that the latent embedding emergently learns to focus on scene aspects pertinent to manipulation tasks. In Figure~\ref{fig:visualization}, we present four distinct images from ScanNet featuring table scenes, a common workspace for robots. We offer a visualization of the final embedding from the ResNet18 pretrained with DPR. Within these examples, the latent embedding of the pretrained model notably gravitates towards the table area, aligning well with the representation observed in the depth map. This observation underscores the efficacy of DPR for manipulation tasks, attributed in part to its innate concentration on task-relevant zones.

\subsection{Real-world Experiments} 
We demonstrate that the pretrained vision backbone works well in practical environments. In our real-world experiment, we use a Franka Emika Panda equipped with a parallel jaw gripper and two RealSense D435i cameras (i.e., eye-to-hand and eye-in-hand) to capture the RGB images. Figure~\ref{fig:real-world} shows a snapshot of the real-world setup. More details are shown in the supplementary video. 

\begin{figure}[t]
\centering
\includegraphics[width=\linewidth]{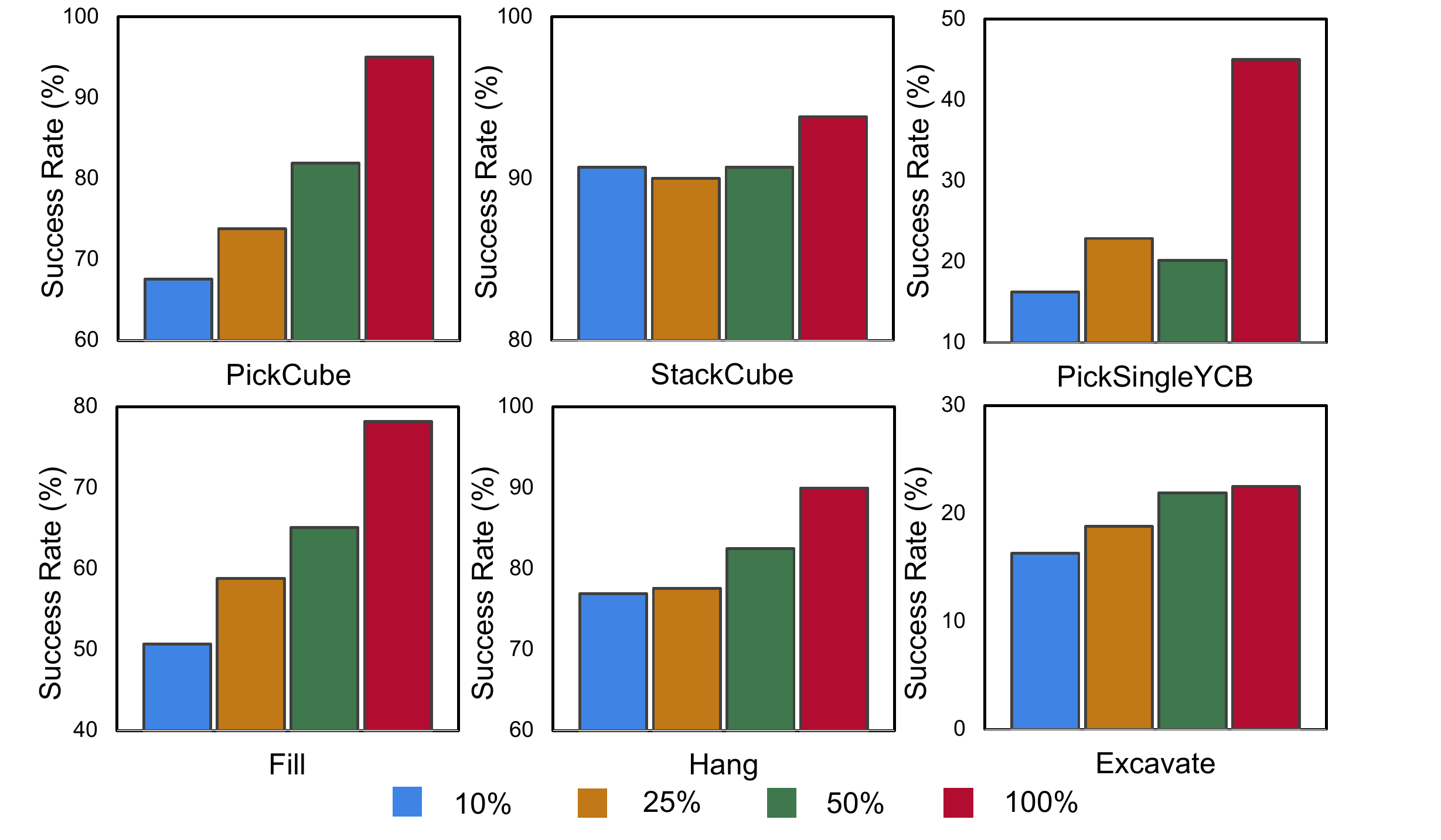}
\caption{Impact of pretraining data size. Pretraining the visual backbone using a large-scale dataset indeed enhances performance.}
\label{fig:size}
\end{figure}

\begin{figure}[t]
\centering
\includegraphics[width=0.85\linewidth]{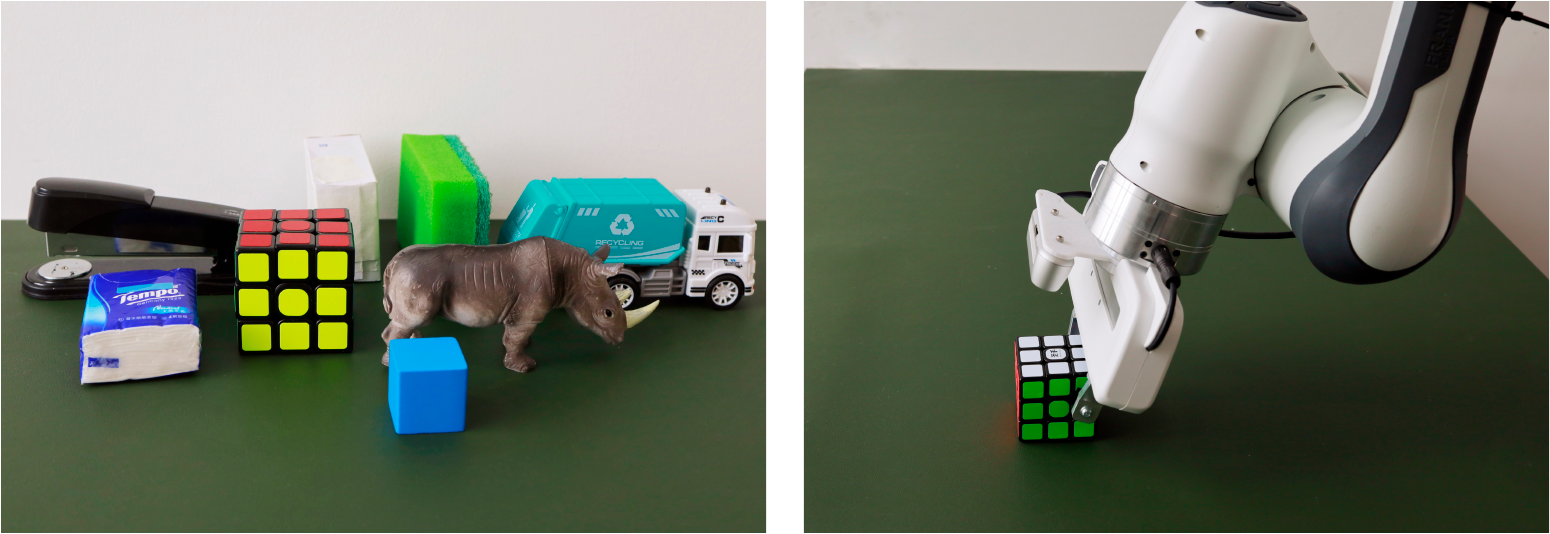}
\caption{A real-world experimental setup for object manipulation.}
\label{fig:real-world}
\end{figure}

\begin{figure}[t]
\centering\includegraphics[width=0.85\linewidth]{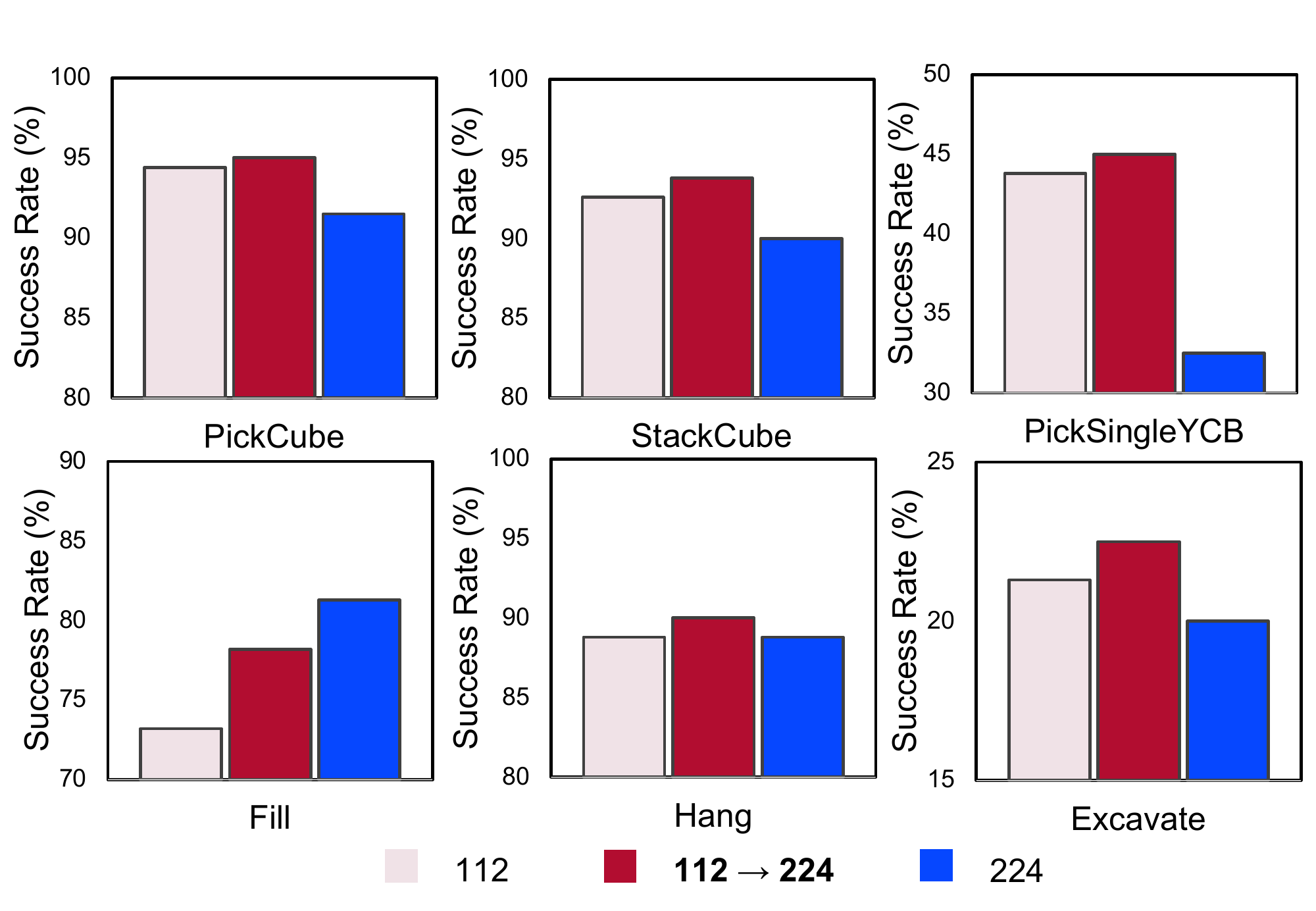}
\caption{Effectiveness of Resolution Adaptation. We observe that adapting the resolution from 112 to 224 performs better than fixed 112. By contrast, using a fixed resolution 224 presents a drop in performance, presumably due to the overfitting in the pretraining stage.}
\label{fig:resolution}
\end{figure}

\subsection{Ablation Study}

\noindent
\textbf{Effectiveness of proprioception injection.} We demonstrate in Table~\ref{table:state} that the proprioception injection is crucial to manipulation tasks. We report the average success rate on Maniskill2 over six tasks. Our proprioception injection method is evaluated on three pretrained weights, i.e., MOCO~\cite{moco}, Pri3D~\cite{pri3d}, and our proposed DPR. it is obvious that ours with proprioceptive fusion strategy consistently performs better than methods that do not have state information. 

\noindent
\textbf{How does the generalization vary with scales of the pretrain dataset?} Our main experiments leverage 240K indoor scene data in ScanNet to perform the pretraining. Here, we investigate whether the size of the data matters. We evaluate 10\%, 25\%, and 50\% of the total training data and see whether reducing the size of pretraining data affects the manipulation success rate. As shown in Figure \ref{fig:size}, increasing the pretraining data size progressively improves the downstream manipulation tasks. Also, learning with the visual backbone pretrained with the whole dataset consistently outperforms that with partial data. 

\noindent
\textbf{Resolution adaptation.}  We measure the impact of changing the resolution during the pretraining on the performance of downstream robot manipulation tasks. We compare models that are trained from scratch using a fixed resolution of either 112 or 224 for 50 epochs with our proposed resolution adaptation techniques. In Figure~\ref{fig:resolution}, we report the success rate of each setting. We can observe that the resolution adaptation method achieves greater performance gain than training on a fixed 112 resolution. Surprisingly, high-resolution training presents a drop in performance, presumably due to overfitting. It is worth mentioning that training with high resolution comes at a high cost: training at 224 is approximately 24$\%$ more compute-intensive than training at 112.

\section{Conclusion}
This work investigates how to pretrain the vision backbone on 3D data and transfer the pretrained visual representation to the robotic manipulation tasks. This is driven by the fact that handling 3D data is time-intensive, and the equipment required to capture this information can be costly. In response, we introduce a depth-aware framework that employs a depth map as supplementary data. This aids contrastive learning by extracting visual features in a self-supervised fashion. We present a proprioception injection method to fuse proprioception data with policy models. Collectively, our methodology showcases robust and adaptable proficiency of DPR across a variety of manipulation tasks applicable to both simulated and real robots.
\clearpage
\bibliographystyle{IEEEtran}
\bibliography{main}

\end{document}